%% file: main.tex
\documentclass[runningheads]{llncs}

\usepackage[T1]{fontenc}

\input{preamble}
\usepackage{cite}
\usepackage[pagebackref,breaklinks,colorlinks]{hyperref}
\hypersetup{colorlinks, citecolor=blue}
\usepackage{tikz}
\usepackage{wrapfig}
\usepackage[capitalize]{cleveref}
\crefname{section}{Sec.}{Secs.}
\Crefname{section}{Section}{Sections}
\Crefname{table}{Table}{Tables}
\crefname{table}{Tab.}{Tabs.}
\newcommand{\repeatthanks}{\textsuperscript{\thefootnote}}
\usepackage[width=122mm,left=12mm,paperwidth=146mm,height=193mm,top=12mm,paperheight=217mm]{geometry}

\usepackage{gensymb}
\usepackage{orcidlink}

\usepackage{stackengine}

\begin{document}

\title{SFDLA: Source-Free Document Layout Analysis}

\titlerunning{SFDLA}

\author{
Sebastian Tewes\,\orcidlink{0009-0000-0418-0131}\inst{1,\thanks{Equal contribution. \text{\textdagger} Corresponding author.}} \and
Yufan Chen\,\orcidlink{0009-0008-3670-4567}\inst{1,\repeatthanks} \and
Omar Moured\,\orcidlink{0000-0003-4227-8417}\inst{1} \and  
Jiaming Zhang\,\orcidlink{0000-0003-3471-328X}\inst{1,2,\text{\textdagger}} \and \\ 
Rainer Stiefelhagen\,\orcidlink{0000-0001-8046-4945}\inst{1}}

\authorrunning{S. Tewes et al.}
\institute{CV:HCI lab, Karlsruhe Institute of Technology, Germany. \and
CVG, ETH Zurich, Switzerland. 
}

\maketitle
\begin{figure}
    \centering
    \vskip -6ex
    \includegraphics[width=0.99\linewidth]{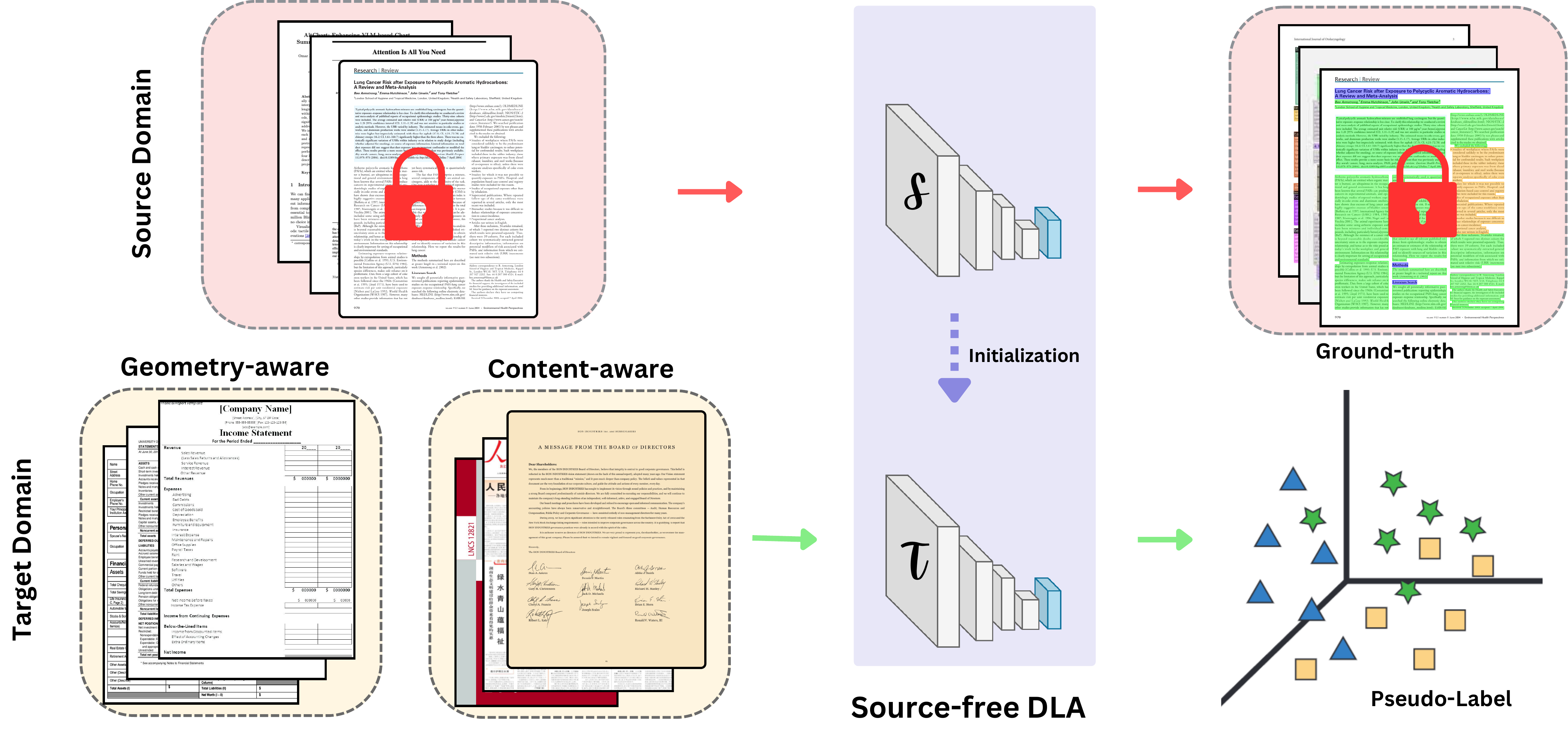}
    \vskip -2ex
    \caption{Overview of Source-Free Document Layout Analysis (SFDLA), adapting the source-domain fine-tuned model (\eg, \(\mathcal{S}\) for scientific papers) to the target domain (\eg, \(\mathcal{T}\) for financial reports), without access to any source data or target labels. 
    }
    \vskip -8ex
    \label{fig:sfdla-overview}
\end{figure}

\input{chapters/abstract}

\input{chapters/introduction}

\input{chapters/relatedwork}

\input{chapters/method}

\input{chapters/experiment}

\input{chapters/conclusion_and_future_work}

\bibliographystyle{splncs04}
\bibliography{main}

\end{document}

%% file: preamble.tex
\usepackage[dvipsnames, table]{xcolor}

\usepackage{diagbox}
\usepackage{array}
\usepackage{mathrsfs}
\usepackage{makecell}
\usepackage{pifont}
\usepackage{placeins}
\usepackage[linesnumbered,ruled,vlined]{algorithm2e}

\usepackage{graphicx}
\usepackage{amsmath}
\usepackage{amssymb}
\usepackage{booktabs}
\usepackage{mathtools}
\usepackage[accsupp]{axessibility}
\usepackage{color}
\usepackage{array}
\usepackage{pdflscape}
\usepackage[longtable]{multirow}
\usepackage{longtable}
\usepackage{tabularray}
\usepackage{booktabs}
\usepackage{times}
\usepackage{epsfig}
\usepackage{amsmath}
\usepackage{bbm}
\DeclareMathAlphabet\mathbfcal{OMS}{cmsy}{b}{n}
\usepackage{makecell}
\usepackage{siunitx}
\usepackage{svg}
\usepackage{float}
\usepackage{comment}
\usepackage{lipsum}
\usepackage{stfloats}
\usepackage{multirow}
\usepackage{bm}
\usepackage{etoolbox}
\usepackage{icomma}
\usepackage{array}
\usepackage{tabulary}
\usepackage{xcolor}
\usepackage{paralist}
\usepackage{booktabs}
\usepackage{adjustbox}
\usepackage{caption}
\usepackage{arydshln}
\usepackage{rotating}

\definecolor{gray}{rgb}{0.3,0.3,0.3}
\definecolor{blue}{rgb}{0,0.5,1}
\definecolor{mask_red}{rgb}{1,0,0.8}
\definecolor{green}{rgb}{0.2,1,0.2}
\definecolor{rblue}{rgb}{0,0,1}
\definecolor{lightblue}{HTML}{6495ed}
\definecolor{lightred}{HTML}{F19C99}

\definecolor{graytablerow}{gray}{0.6}

\usepackage{pifont}

\usepackage{tabu}
\usepackage[pro]{fontawesome5}

\usepackage{tikz}
\newcommand*\circled[1]{\tikz[baseline=(char.base)]{
\node[shape=circle,fill=gray,inner sep=0.5pt] (char) {\textcolor{white}{\small \textbf{#1}}};}}

\makeatletter
\DeclareRobustCommand\onedot{\futurelet\@let@token\@onedot}
\def\@onedot{\ifx\@let@token.\else.\null\fi\xspace}

\def\eg{\emph{e.g}\onedot} 
\def\ie{\emph{i.e}\onedot}

\makeatother

%% file: chapters/abstract.tex
\begin{abstract}
Document Layout Analysis (DLA) is a fundamental task in document understanding. However, existing DLA and adaptation methods often require access to large-scale source data and target labels. This requirements severely limiting their real-world applicability, particularly in privacy-sensitive and resource-constrained domains, such as financial statements, medical records, and proprietary business documents. According to our observation, directly transferring source-domain fine-tuned models on target domains often results in a significant performance drop (Avg. ${-}32.64\%$). 
In this work, we introduce \textbf{Source-Free Document Layout Analysis (SFDLA)}, aiming for adapting a pre-trained source DLA models to an unlabeled target domain, without access to any source data. To address this challenge, we establish the first SFDLA benchmark, covering three major DLA datasets for geometric- and content-aware adaptation. Furthermore, we propose \textbf{Document Layout Analysis Adapter (DLAdapter)}, a novel framework that is designed to improve source-free adaptation across document domains. 
Our method achieves a ${+}4.21\%$ improvement over the source-only baseline and a ${+}2.26\%$ gain over existing source-free methods from PubLayNet to DocLayNet.
We believe this work will inspire the DLA community to further investigate source-free document understanding. To support future research of the community, the benchmark, models, and code will be publicly available at \url{https://github.com/s3setewe/sfdla-DLAdapter}.

\end{abstract}

%% file: chapters/introduction.tex
\section{Introduction}
\label{sec:intro}

DLA focuses on identifying and segmenting structural elements within a document, such as text, images, and tables. This process converts unstructured documents into machine-readable formats, improving digitization, automation, and data extraction.
A key challenge in DLA is the heterogeneity of document structures across domains. The common practice for adapting models to these layouts is supervised fine-tuning using annotated data. However, in numerous practical applications, accessing source domain data may not be feasible due to concerns about personal privacy, confidentiality, and copyright \cite{bhatt2021survey}.

To mitigate the reliance of domain adaptation methods on source data, the Source-Free Domain Adaptation (SFDA) paradigm \cite{10452835} has been introduced in recent years. This approach has gained significant attention and has been studied in various fields, including image classification \cite{kundu2020universal,xia2021adaptive}, semantic segmentation \cite{liu2021source}, and object detection\cite{Li2020AFL,xiong2022source}.
SFDA provides a compelling solution to the challenges posed by domain shifts. Unlike traditional supervised learning, which assumes identical distributions between source and target domains and relies on concurrent access to both, SFDA applies a fully self-supervised adaptation phase, refining the model without requiring labeled target data. It follows a two-stage process as shown in Figure \ref{fig:sfdla-overview}: first, learning discriminative feature representations from the source domain in a fully supervised manner, which could be on a similar task dataset (e.g., scientific document layout), and then leveraging self-supervised learning mechanisms to progressively align the model to the target domain (e.g., private medical or financial documents). This adaptation process enables the model to capture domain-specific variations, mitigate distribution discrepancies, and enhance generalization without explicit supervision.

\begin{figure}[!h]
    \centering
    \includegraphics[width=\linewidth]{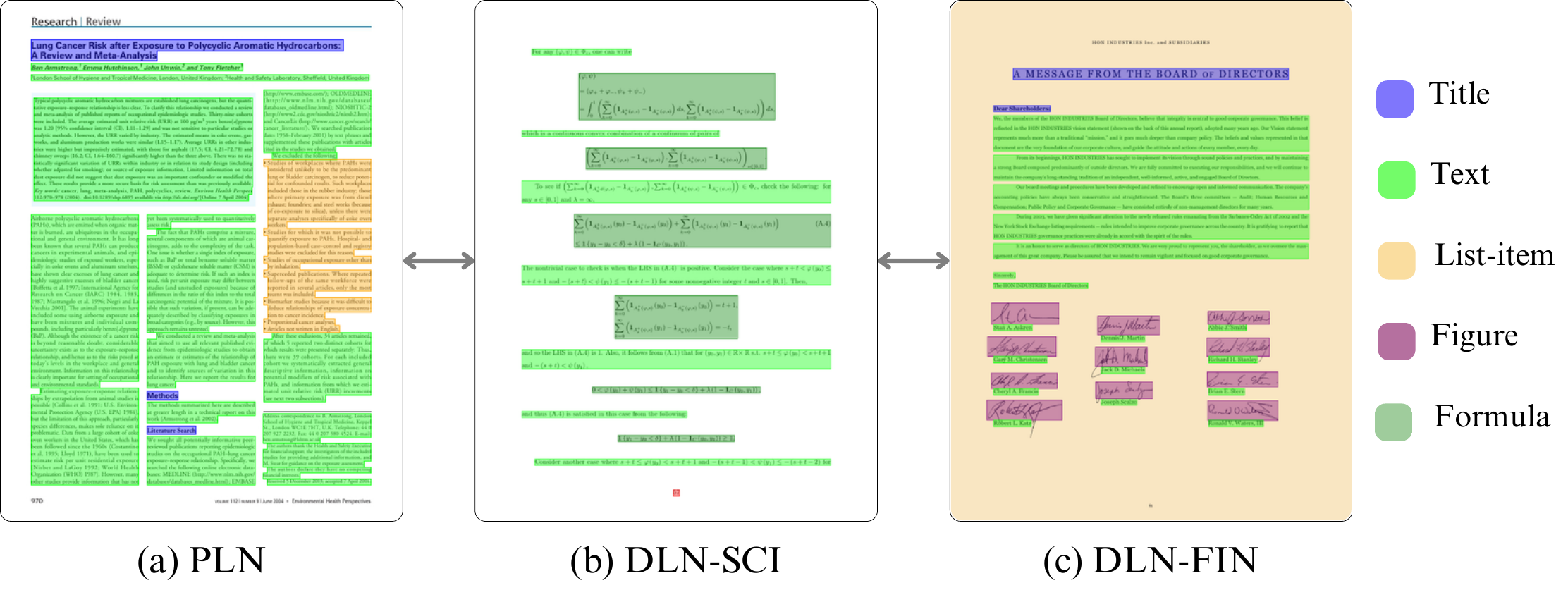}
    \caption{Sample documents with ground-truth annotations. (a)-(b) are STEM documents (biology, math), while (c) is a financial document. Color code denote different layout elements. The structural diversity highlights the need for adaptive DLA across domains.}
    \label{fig:sfdla-samples}
\end{figure}

Because, to the best of our knowledge, there is no existing benchmark for the Source-Free Document Layout Analysis (SFDLA), we introduce the SFDLA challenge in this work. Figure \ref{fig:sfdla-samples} showcases three documents sampled from the PubLayNet \cite{zhong2019publaynet} and DocLayNet \cite{10.1145/3534678.3539043} datasets to illustrate the challenges involved. Figures \ref{fig:sfdla-samples}-(a) \& (b) show documents from the STEM domain; however, (a) from PubLayNet follows a two-column layout with dense text and list elements, while (b) adopts a single-column format with dense mathematical expressions. This challenge extends to labeling guidelines, as seen in (c), where a financial document shares a similar layout but treats signatures, absent in scientific documents, as image elements. Considering \circled{1} the time-intensive annotation process, \circled{2} privacy concerns, and \circled{3} the complexity of establishing a unified layout representation standard, we are motivated to introduce this benchmark.

However, SFDA presents its own set of challenges \cite{li2024comprehensive}. Since adaptation relies solely on the target domain's unlabeled data, the model must infer and exploit its intrinsic structures without guidance from labeled samples. This can lead to pseudo-labeling inconsistencies, where erroneous predictions reinforce incorrect feature alignment, potentially hampering convergence. Moreover, the lack of direct supervision increases sensitivity to domain complexity, especially in scenarios with substantial inter-domain discrepancies. Addressing these challenges requires robust self-supervised regularization techniques that enhance feature stability and maintain semantic consistency across domains.

We address these by introducing DLAdapter, a dual-teacher framework to balance semantic learning and adaptation. It employs a static teacher, updated via Exponential Moving Average (EMA) for stable feature extraction, and a dynamic teacher, which adapts to recent student updates for flexible supervision. By doing so, we demonstrate that such guided representation learning enhances representations for the target data.

\noindent In summary, this paper makes the following contributions:
\begin{itemize}
    \renewcommand{\labelitemi}{\textbullet}
    \setlength{\itemsep}{6pt}
    \item We explore source-free domain adaptation in document layout analysis, identifying critical domain shifts and their impact on adaptation performance.  
    \item We establish the first benchmark for source-free document layout adaptation, paving the way for privacy-compliant, real-world document understanding applications.  
    \item We develop DLAdapter, a state-of-the-art dual-teacher method that enhances pseudo-label reliability and feature alignment, enabling cross-domain learning.  
\end{itemize}

%% file: chapters/relatedwork.tex
\section{Related Work}
\label{sec:related_work}

\noindent\textbf{Document Layout Analysis.} Analyzing document layout is a fundamental task in document understanding. The introduction of large-scale document layout benchmarks, such as M$^6$Doc~\cite{cheng2023M$^6$Doc}, PubLayNet(PLN)~\cite{zhong2019publaynet}, and DocLayNet(DLN)~\cite{10.1145/3534678.3539043}, has significantly advanced the field by unifying layout representations across diverse domains. For instance, DocLayNet categorizes documents into six distinct types, including \textit{financial reports}, \textit{scientific articles}, and \textit{law documents}. These benchmarks have led to the development of several SOTA models \cite{chen2024rodla,10.1145/3503161.3547911,huang2022layoutlmv3,bhowmik2023document}.
Recently, RoDLA \cite{chen2024rodla} has addressed the layout robustness challenge by applying self-attention across the channel dimension, effectively aggregating local features and emphasizing spatially relevant tokens, resulting in improved resilience to various document perturbations.
CNN-based object detectors, such as the YOLO family~\cite{deng2024yolo} and Faster R-CNN~\cite{NIPS2015_14bfa6bb}, continue to perform well in layout detection tasks under fully supervised training.
Beyond single-modality approaches, VGT \cite{da2023vision} employs a two-stream architecture that integrates visual features from a vision transformer and textual information processed by a pre-trained grid transformer. This design captures both token-level and segment-level semantics.

Nevertheless, fully supervised methods fail to generalize to unseen layouts and require large amounts of annotated data for fine-tuning across varying layouts. Our preliminary experiments with Faster R-CNN (Table \ref{tab:train_inference_map50}) reveal that training with a single domain results in a significant drop in mAP of 32.64\% when tested with other domains.

\begin{table}[t]
    \centering
    \caption{Cross-domain evaluation of Faster R-CNN \cite{NIPS2015_14bfa6bb} on DocLayNet \cite{10.1145/3534678.3539043}. The model is trained in a single-category setting and tested on the validation set’s unseen categories. The table reports mAP-50 for Scientific Articles (Sci.), Financial Reports (Fin.), Manuals (Man.), and Laws \& Regulations (Laws). Red values representing the relative performance drop from the trained category.}
    \renewcommand{\arraystretch}{1.2}
    \setlength{\tabcolsep}{8pt}
    \resizebox{0.99\textwidth}{!}{
    \begin{tabular}{c|cccc}
        \toprule
        \diagbox[innerwidth=6em, height=2.5em, linewidth=0.5pt]{\textbf{Source}}{\textbf{Target}} 
        & \textbf{Sci.} & \textbf{Fin.} & \textbf{Man.} & \textbf{Laws} \\ 
        \midrule
        \textbf{Sci.}  & \cellcolor{gray!20} 84.32 & 18.72 \textcolor{red}{(-65.60\%)} & 36.77 \textcolor{red}{(-47.55\%)} & 42.24 \textcolor{red}{(-42.08\%)} \\ 
        \textbf{Fin.}  & 38.74 \textcolor{red}{(-14.58\%)} & \cellcolor{gray!20} 53.32 & 45.93 \textcolor{red}{(-7.39\%)} & 43.69 \textcolor{red}{(-9.63\%)} \\ 
        \textbf{Man.}  & 21.59 \textcolor{red}{(-39.92\%)} & 22.34 \textcolor{red}{(-39.17\%)} & \cellcolor{gray!20} 61.51 & 46.99 \textcolor{red}{(-14.52\%)} \\ 
        \textbf{Laws}  & 31.91 \textcolor{red}{(-39.59\%)} & 22.08 \textcolor{red}{(-49.42\%)} & 49.33 \textcolor{red}{(-22.17\%)} & \cellcolor{gray!20} 71.50 \\ 
        \bottomrule
    \end{tabular}}
    \vskip -2ex
    \label{tab:train_inference_map50}
\end{table}

\noindent\textbf{Source-Free Domain Adaption.} A central challenge in machine learning is generalization—ensuring that a model performs well on unseen data. In real-world applications, models often encounter distribution shifts, leading to performance degradation. Domain adaptation mitigates this issue by compensating for these differences. Traditional methods, such as Unsupervised Domain Adaptation (UDA) \cite{chen2022learningdomainadaptiveobject, mohamadi2024featurebasedmethodsdomain, li2024daadalearningdomainawareadapter, li2020cross}, assume full access to the target domain data during adaptation. However, in many real-world scenarios, this assumption is impractical due to the reasons discussed earlier. SFDA relaxes this requirement, assuming that only the trained source model is available without direct access to the target domain data. While SFDA has been widely explored for classification tasks \cite{9157725, XIAO2024110533, cheng2024towards, 9528982, kothandaraman2023salad}, its extension to object detection, known as Source-Free Object Detection (SFOD), remains relatively underexplored. SFOD typically relies on mean-teacher self-training, where a student-teacher framework generates pseudo-labels \cite{10.5555/3294771.3294885}.

The first SFOD framework, introduced in 2020, proposed a self-training approach where a source-trained network generates pseudo-labels to optimize performance on the target domain \cite{Li2020AFL}. Since then, several studies have enhanced SFOD by improving pseudo-labeling quality, employing style augmentation \cite{9878540}, adversarial alignment \cite{10.1609/aaai.v37i1.25119}, instance relation graphs \cite{10204606}, adaptive thresholds \cite{10096635, 10.1145/3581783.3612273}, and dynamic learning processes \cite{10377810, khanh2024dynamicretrainingupdatingmeanteacher}. 
A recent notable approach we investigate is Instance Relation Graph (IRG) \cite{vs2023instance}, which enhances feature alignment through contrastive representation learning by modeling relationships between object proposals. However, IRG lacks mechanisms for training stability and dynamic pseudo-label refinement, making it prone to error accumulation and confirmation bias, where early mispredictions reinforce themselves over time.

%% file: chapters/method.tex
\section{Source-Free Document Layout Analysis}
DLA is a fundamental step for document information retrieval and analysis. However, a persistent challenge for DLA is the heterogeneity of document structures and styling across different domains. Business invoices, scientific papers, and financial reports follow significantly different layouts, typographical conventions, and label guidelines. Consequently, a DLA model trained on one domain tends to degrade in performance when directly applied to another due to the domain shift~\cite{yoon2024enhancingsourcefreedomainadaptive}.

\begin{figure}[!ht]
    \centering
    \includegraphics[width=\linewidth]{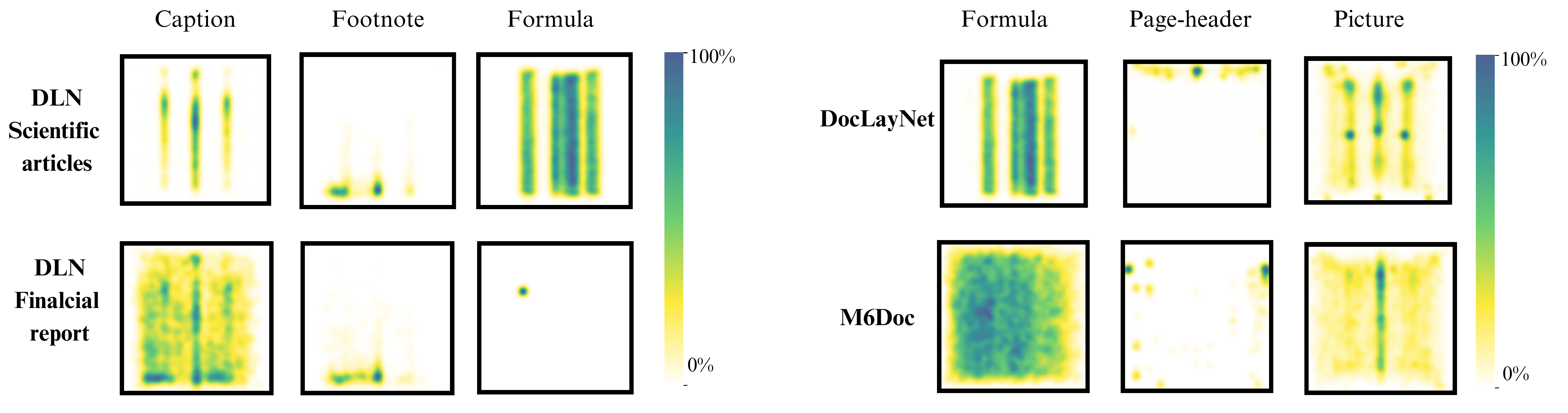}
    \caption{Visualization of distribution of layout elements in document images. Left: Domain gap due to content difference. Right: Domain gap due to geometric difference.}
    \label{fig:vis_analysis}
\end{figure}

Fig.~\ref{fig:vis_analysis} illustrates diverse layout distributions within different document types and datasets. A scientific paper often adopts a multi-column format with charts and images, meanwhile, a financial record may include signatures, which are treated as \emph{images} category in datasets that do not appear in academic articles. This variability causes difficulty for a single model to generalize seamlessly across domains. A common strategy to handle domain distinctions is fine-tuning or Unsupervised Domain Adaptation (UDA) methods, which typically rely on simultaneous access to source and target data. Nevertheless, in realistic scenarios, source-domain data cannot be transferred due to privacy or confidentiality constraints, \eg, sensitive data in healthcare or finance.

\begin{figure}[!h]
    \centering
    \includegraphics[width=\linewidth]{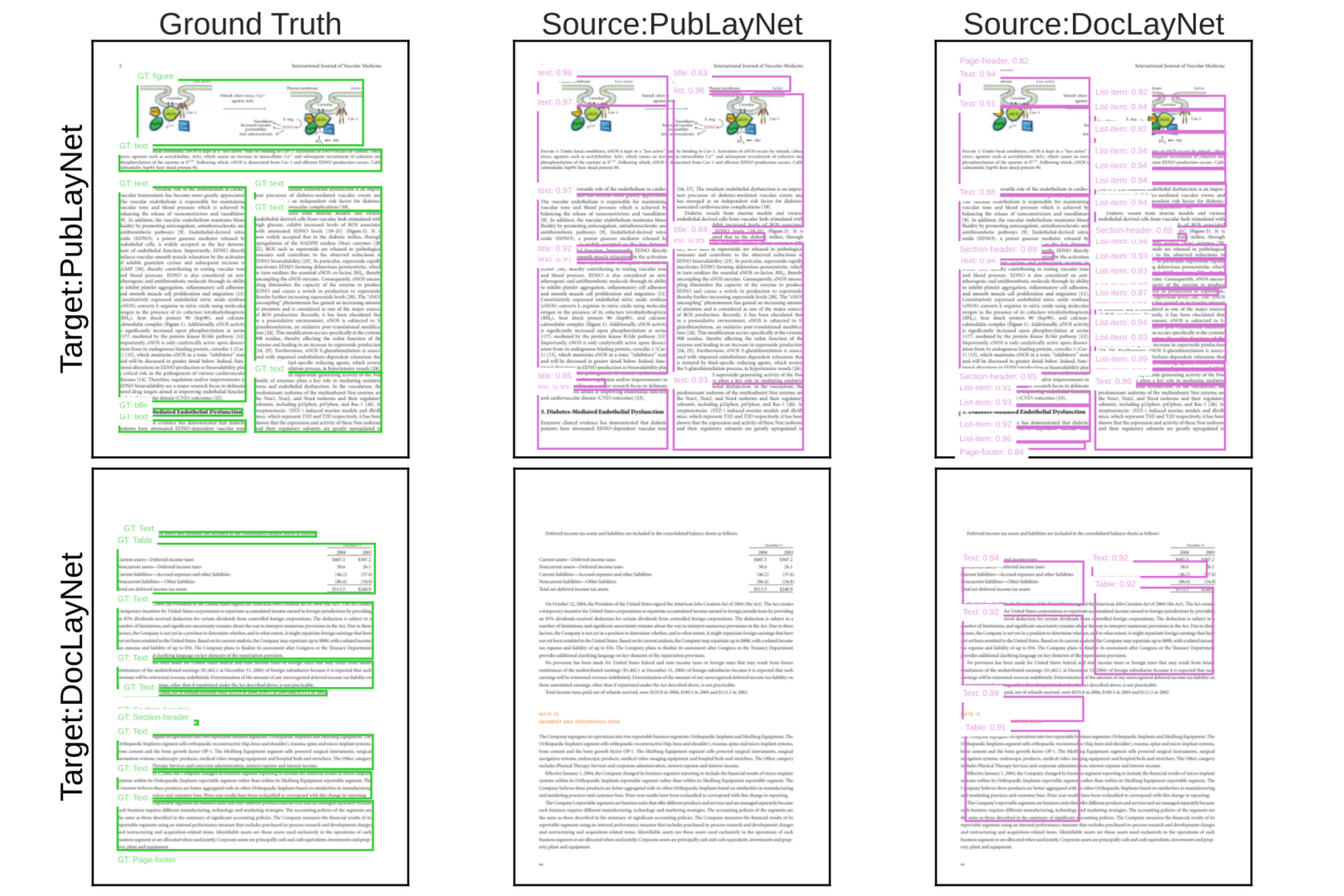}
    \caption{Cross-domain Faster R-CNN~\cite{NIPS2015_14bfa6bb} performance visualization between PubLayNet~\cite{zhong2019publaynet} and DocLayNet~\cite{10.1145/3534678.3539043} without any adaptation.}
    \label{fig:vis_bench}
\end{figure}

Under such constraints, simply deploying a source-trained model on a new domain can lead to drastic performance drops. \textbf{Source-Free Domain Adaptation (SFDA)}  has recently emerged as a compelling solution for scenarios where the source data are not available during adaptation~\cite{9157725, XIAO2024110533, cheng2024towards, 9528982}. In SFDA, a pre-trained source model is provided, but no source examples can be used. Adaptation must consequently rely on unlabeled target-domain data, ensuring privacy and reducing data-transfer overhead. SFDA has merely been explored for typical vision tasks such as classification and semantic segmentation. We define the \textbf{Source-Free Document Layout Analysis (SFDLA)} task, which focuses specifically on adapting DLA models without source-domain data. We establish a novel benchmark comprising different standard datasets, i.e., PubLayNet~\cite{zhong2019publaynet}, DocLayNet~\cite{10.1145/3534678.3539043}, and M$^6$Doc~\cite{cheng2023M$^6$Doc}, for cross-domain document layout without source data, as shown in Fig.~\ref{fig:vis_bench}. Our analysis shows that naive application of a source model leads to significant performance loss, and adaptation proves challenging due to characteristic annotation policies, specific layouts, and granularity variations.

\section{DLAdapter}
\label{sec:DLAdapter}

We propose the novel \textbf{DLAdapter} framework to enable the SFDLA task. DLAdapter builds on a standard object detection architecture that has been trained on source-domain documents. Given only the pre-trained weights and unlabeled target-domain images, DLAdapter performs self-supervised adaptation without accessing source data. Fig.~\ref{fig:CADSFDADLA} presents the core components of DLAdapter:

\begin{figure*}[!ht]
    \centering
    \includegraphics[width=\textwidth]{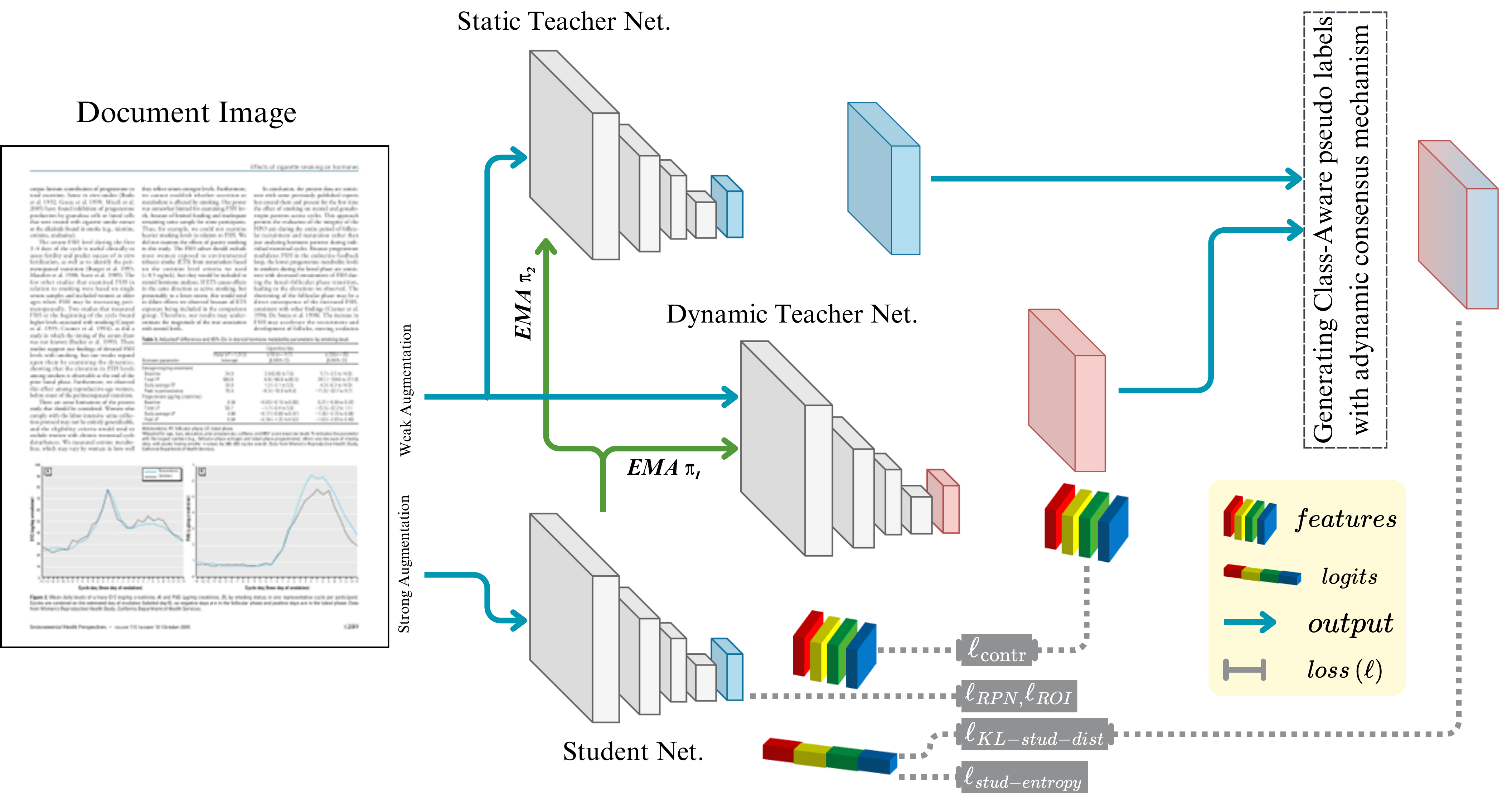}
    \caption{Architecture of our method DLAdapter for SFDLA task: We follow a student-teacher framework with a static and dynamic teacher that generates class-aware pseudo-labels for the student in a consensus mechanism. Various supporting losses (yellow) let the student model learn, which in turn updates the two teacher models via EMA.}
    \label{fig:CADSFDADLA}
\end{figure*}

\begin{itemize}
  \setlength\itemsep{1em}
    \item \textbf{Student Model}: An initial source-trained network iteratively adapted to the target domain based on pseudo-labels derived from dual-teacher predictions.
    \item \textbf{Dual-Teacher Models (Static \& Dynamic)}: Two teacher networks each maintain an exponential moving average (EMA) of the student’s parameters but at different update rates. The static teacher retains stable, long-term source knowledge, while the dynamic teacher speedily captures target-specific features. Their synergy mitigates overfitting to noisy targets and catastrophic forgetting of source information.
    \item \textbf{Consensus Pseudo-Labeling}: The two teachers' predictions are fused for high-confidence pseudo-labels. Matching detections become consensus labels with elevated confidence, whereas unmatched or inconsistent predictions are down-weighted. These pseudo-labels supervise the student model in a self-supervised manner.
\end{itemize}

\subsection{Student Model with Dual-Teacher Framework}
\label{sec:dual_teacher}
A major difficulty in SFDLA is the overfitting to noisy pseudo-labels. To address this issue, we draw inspiration from PETS~\cite{liu2023PETS} and propose a dual-teacher framework comprising a static teacher and a dynamic teacher. Let \(\theta_{\mathrm{teac}}\) and \(\theta_{\mathrm{stud}}\) denote the teacher and student weights, respectively. After every update interval, we compute:
\begin{equation}
    \label{eq:teacher_ema}
    \theta_{\mathrm{teac}} 
    \;\leftarrow\;
    \pi\,\theta_{\mathrm{teac}} 
    \;+\; 
    \bigl(1-\pi\bigr)\,\theta_{\mathrm{stud}},
\end{equation}
where $\pi \in [0,1]$ is the momentum factor, $\bigl(1-\pi\bigr)$ determines the portion of updated parameters incorporated at each update.

\noindent \textbf{Dynamic Teacher Model.} As a normal single teacher framework, the dynamic teacher employs a larger \(\pi_{\mathrm{1}}\), while the dynamic teacher model updates every 2000 iterations in training. This allows the dynamic teacher to capture the student’s most recent learning in near real time, reflecting subtle target-domain cues as soon as they emerge. Moreover, since each update only partially shifts its parameters, the dynamic teacher avoids making wrong weights shift with student errors or noisy pseudo-labels.

\noindent \textbf{Static Teacher Model.} By contrast, we choose a relatively small \(\pi_{\mathrm{2}}\), but still large enough to prevent noisy weights shifting in the static teacher model. After each training epoch, the static teacher absorbs a greater fraction of the student’s current weights, resulting in a sizeable parameter shift. Infrequent updates shield it from transient gradient noise, accumulating the student’s progress more conservatively over a larger training interval. Meanwhile, each static update assimilates a substantial portion of the student’s parameters, this teacher can also serve as a robust reference if rapid self-training leads to severe label drift, which effectively stabilizes the student model.

Together, this design provides the student with broad, stable guidance on discovered patterns, helping ensure more reliable convergence on the target domain. The whole DLAdapter working flow is shown as in Algorithm~\ref{alg:CAD-SFDA-DLA}. \input{code/DLAdapter}

In Algorithm~\ref{alg:CAD-SFDA-DLA}, $\hat{b}, \hat{c}, \hat{s}, \hat{l}$ represent the components of the predicted bounding box, class category, confidence score, and label distribution from the teachers, respectively. The notation $x_{\mathit{weak}}$ and $x_{\mathit{strong}}$ indicates the weakly and strongly augmented input image $x$. Besides, $F$ denotes extracted feature maps.

\subsection{Consensus Pseudo-Labeling}
\label{sec:consensus_pseudo_label}
After dual teacher prediction, we merge these outputs into a refined set of pseudo-labels. First, bounding boxes with sufficiently high IoU and consistent class predictions across both teachers are matched and assigned a high confidence score, forming consensus detections. Any predictions that do not find verification from the other teacher are treated as mismatched and thus have low confidence to reflect uncertainty. Low-confidence predictions are filtered out by threshold, and redundant detections are removed via non-maximum suppression (NMS) operations. These consensus pseudo-labels then serve as surrogate ground truths for the student’s training losses. Compared to single-teacher self-training, this consensus-based approach significantly reduces noise and error propagation with more reliable supervision. 

By leveraging dual-teacher networks to balance stability and adaptability, DLAdapter generates comparably high-quality pseudo-labels for the unlabeled target data, enabling it to outperform zero-shot transfer and achieve state-of-the-art performance in SFDLA.

%% file: code/DLAdapter.tex
\vspace{-5ex}
\begin{algorithm}[!h]
\SetAlgoLined
\KwIn{Pre-trained source model $f_{\theta_{D_S}}$, unlabeled target domain data $D_{T}$}
\KwOut{Adapted student model $f_{\theta_{stud}}$}

\textbf{Initialization:} 
\[
\quad f_{\theta_{t,dyn}}, \; f_{\theta_{t,stat}}, \; f_{\theta_{stud}} \;\gets\; f_{\theta_{D_S}}
\]
\For(\tcp*[h]{outer loop over epochs}){$epoch \gets 1$ \KwTo $n_{epochs}$}{
    \For(\tcp*[h]{inner loop over target samples}){$x_{target}^i \in D_{T}$}{
        
        \tcc{\footnotesize \textbf{(1) Static Teacher Inference}}
        $r_{t,stat}^i = (\hat{b}_{t,stat}^i,\;\hat{c}_{t,stat}^i,\;\hat{s}_{t,stat}^i,\;\hat{l}_{t,stat}^i)
          \gets f_{\theta_{t,stat}}^{\text{inference}}\bigl((x_{target}^i)_{\text{weak}}\bigr)$
        
        \tcc{\footnotesize \textbf{(2) Dynamic Teacher Inference}}
        $r_{t,dyn}^i = (\hat{b}_{t,dyn}^i,\;\hat{c}_{t,dyn}^i,\;\hat{s}_{t,dyn}^i,\;\hat{l}_{t,dyn}^i),\;
        (Ft_{t,dyn}^i)_{\text{weak}} 
          \gets f_{\theta_{t,dyn}}^{\text{inference}}\bigl((x_{target}^i)_{\text{weak}}\bigr)$
        
        \tcc{\footnotesize \textbf{(3) Consensus Mechanism}}
        $\hat{y}_{pseudo}^i = \bigl(\hat{b}^i,\;\hat{c}^i,\;\hat{s}^i,\;\hat{l}^i\bigr) 
          \gets g_{\text{consensus}}\bigl(r_{t,stat}^i,\;r_{t,dyn}^i\bigr)$
        
        \tcc{\footnotesize \textbf{(4) Student Inference}}
        $\bigl((Ft_{s}^i)_{\text{strong}},\;l_{s}^i,\;L_{RPN_s}^i,\;L_{ROI_s}^i\bigr)
           \gets f_{\theta_{stud}}^{\text{train}}\bigl((x_{target}^i)_{\text{strong}},\;\hat{y}_{pseudo}^i\bigr)$\;
        $L_{RPN_s}^i \gets L_{cls_s}^{RPN^i} + L_{reg_s}^{RPN^i}$\;
        $L_{ROI_s}^i \gets L_{cls_s}^{ROI^i} + L_{reg_s}^{ROI^i}$\;
        
        \tcc{\footnotesize \textbf{(5) Self-Training Losses}}
        $L_{\text{KL\_stud\_dis}}^i \;\gets\; \text{soft\_kl\_dist}\bigl(l_s^i,\;\hat{l}^i\bigr)$\;
        $L_{\text{f\_dis}}^i \;\gets\; \text{f\_dis}\bigl((Ft_{t,dyn}^i)_{\text{weak}},\;(Ft_{s}^i)_{\text{strong}}\bigr)$\;
        $L_{\text{stud\_entropy}}^i \;\gets\; \text{entropy}\bigl(l_{s}^i\bigr)$\;
        $L_{\text{cont}}^i \;\gets\; \text{cont}\bigl((Ft_{t,dyn}^i)_{\text{weak}},\;(Ft_{s}^i)_{\text{strong}}\bigr)$\;
        
        \tcc{\footnotesize \textbf{(6) Weighted Loss and Backprop}}
        $factor^i \;\gets\; \bigl(1 + \gamma_{e}\cdot \text{entropy}(\hat{l}_{t,dyn}^i)\bigr)\,\times\,
                              \bigl(1 + \gamma_{p}\cdot \#(\hat{y}_{pseudo}^i)\bigr)$\;
        $\mathscr{L}^i \;\gets\; factor^i \,\times \Bigl[w_{1}\,L_{RPN_s}^i 
               + w_{2}\,L_{ROI_s}^i 
               + w_{3}\,L_{\text{KL\_stud\_dis}}^i 
               + w_{4}\,L_{\text{f\_dis}}^i 
               + w_{5}\,L_{\text{stud\_entropy}}^i
               + w_{6}\,L_{\text{cont}}^i\Bigr]$\;
        
        $\mathscr{L}^i.\text{backward}()$\;
        $f_{\theta_{stud}}.\text{optimizer\_step}()$\;
        
        \If{$i \,\%\; n_{\text{update}} \,=\, 0$}{
            \tcp*[h]{EMA update for dynamic teacher}\;
            $\text{update}\bigl(f_{\theta_{t,dyn}},\, f_{\theta_{stud}},\, rate_{\text{EMA}}^{dyn}\bigr)$
        }
    }
    \tcp*[h]{EMA update for static teacher once per epoch}\;
    $\text{update}\bigl(f_{\theta_{t,stat}},\, f_{\theta_{stud}},\, rate_{\text{EMA}}^{stat}\bigr)$
}
\caption{Pseudo-Code DLAdapter}
\label{alg:CAD-SFDA-DLA}
\end{algorithm}
\vspace{-5ex}

%% file: chapters/experiment.tex
\section{Experiment}
\label{sec:experiments}

\subsection{Implementation Details}
\label{subsec:implementation}
We build our method upon a region-based CNN detector for consistency with prior SFOD works and computational efficiency. In particular, we adopt a Faster R-CNN architecture with a ResNet-50 backbone~\cite{7780459} pre-trained on ImageNet~\cite{5206848}. The source model is first trained on the source domain for 50k iterations with a batch size of 8 across 4 NVIDIA 1080Ti GPUs. The initial learning rate is $0.001$ (with a linear warm-up) and is decayed by a factor of 0.1 at 10k, 20k, and 40k iterations. We then adapt the trained source model to the target domain under the source-free setting, using a batch size of 1 on 2 GPUs. The processing time of our method is 0.198 s per image, the peak memory usage is 1502 MB, and the FLOPS (Floating-point operations per second) is 132.17 G. Our implementation utilizes Detectron2~\cite{wu2019detectron2}. All evaluations use $mAP@0.5$ (mean Average Precision at $50\%$ IoU) as the primary metric.

\subsection{Datasets and Adaptation Benchmarks}
\label{subsec:datasets}

We evaluate DLAdapter on three public DLA datasets, including: PubLayNet~\cite{zhong2019publaynet}, DocLayNet~ \cite{10.1145/3534678.3539043}, and M$^6$Doc~\cite{cheng2023M$^6$Doc}.

\noindent \textbf{PubLayNet}~\cite{zhong2019publaynet} is a large-scale dataset of document layouts automatically derived from PubMed Central articles. It contains 360k page images with about 3.3M annotated objects across five classes, \ie, \textit{text}, \textit{title}, \textit{list}, \textit{figure}, \textit{table}.

\noindent \textbf{DocLayNet}~\cite{10.1145/3534678.3539043} is a diverse, manually-annotated layout dataset covering documents from six categories, including \textit{financial reports}, \textit{scientific articles}, \textit{government tenders}, \textit{manuals}, \textit{patents}, and \textit{laws and regulations}. It comprises 11 semantic layout classes and 80K images, addressing the limitations of low layout variability in previous datasets.

\noindent \textbf{M$^6$Doc}~\cite{cheng2023M$^6$Doc} is a highly diverse collection of 9,080 document pages with 237k annotations. It defines 74 fine-grained layout classes in English and Chinese, making it significantly more challenging due to its broad class spectrum. 

Incompatible label sets complicate cross-dataset evaluation. Each dataset has its own taxonomy and classes with the same names may be defined differently under different annotation guidelines. DocLayNet~\cite{10.1145/3534678.3539043} highlights this issue in their work and recommends evaluating on a reduced common label space for fairness. Following these guidelines, we map classes to a shared subset for each cross-domain experiment. Specifically, for comparison between PubLayNet~\cite{zhong2019publaynet} and DocLayNet~\cite{10.1145/3534678.3539043}, we restrict both to the four common classes, \ie, \textit{figure}, \textit{table}, \textit{text}, and \textit{title}. For the comparison between DocLayNet~\cite{10.1145/3534678.3539043} and M$^6$Doc~\cite{cheng2023M$^6$Doc}, we similarly align them to ten common classes.

Based on these datasets and mappings, we define two type of DLA adaptation settings with five domain adaptation scenarios for our source-free document layout analysis evaluation. All data splitting and evaluation will be made publicly available. 

\begin{itemize}
  \setlength\itemsep{1em}

    \item \textbf{Geometry-aware (Inter-dataset) SFDLA:} 
        \begin{itemize}
            \item PLN$_4$ $\rightarrow$ DLN$_4$: From PubLayNet~\cite{zhong2019publaynet} to DocLayNet~\cite{10.1145/3534678.3539043} on 4 unified classes.
            \item DLN$_{10}$ $\rightarrow$ M$^6$Doc$_{10}$:From DocLayNet~\cite{10.1145/3534678.3539043} to M$^6$Doc~\cite{cheng2023M$^6$Doc} on 10 unified classes.
        \end{itemize}

    \item \textbf{Content-aware (Intra-dataset) SFDLA:} 
        \begin{itemize}
            \item Sci. $\rightarrow$ Fin.: From scientific to financial documents within DocLayNet~\cite{10.1145/3534678.3539043}.
            \item Man. $\rightarrow$ Fin.: From manuals to financial within DocLayNet~\cite{10.1145/3534678.3539043}.
            \item Laws $\rightarrow$ Man.: From laws to manuals within DocLayNet~\cite{10.1145/3534678.3539043}.
        \end{itemize}
\end{itemize}

\subsection{Quantitative Results} 
\label{subsec:quantitativeresults}

\subsubsection{Geometry-aware (Inter-dataset) Results.}
\label{subsubsec:interdataset}
Table~\ref{tab:pln4_dln4_style} reports results for the four-class domain adaptation from PubLayNet~\cite{zhong2019publaynet} to DocLayNet~\cite{10.1145/3534678.3539043}. Our method outperforms the source-only baseline by \textbf{$4.21\% $} and IRG~\cite{10204606} by \textbf{$2.26\%$} at mAP$_{50}$ performance. Notably, we achieve consistent improvements across across all classes except the class \textit{table}. Especially for results with class \textit{text}, our DLAdapter method rises around \textbf{$11\%$} and around \textbf{$8\%$} compared to the source-only and IRG method, respectively, highlighting the effectiveness of our proposed self-training with dual teachers.

\vspace{-4ex}

\begin{table*}[!ht]
    \centering
    \footnotesize
    \caption{Per-class AP and overall mAP$_{50}$ performance in domain adaptation scenario: \textbf{PLN$_4$} $\rightarrow$ \textbf{DLN$_4$}. \textbf{Oracle} is the upper bound using full target labels, \textbf{Source Only} means directly transferring the source-trained model without adaptation, \textbf{IRG}~\cite{10204606} is a baseline for source free domain adaptation, and \textbf{Ours} is the proposed DLAapter method.}
    \label{tab:pln4_dln4_style}
    \resizebox{0.98\textwidth}{!}{
        \setlength{\tabcolsep}{3mm}
        \begin{tabular}{l|l|cccc|c}
        \toprule
        \textbf{Scenario}&\textbf{Method}& \textbf{Figure} & \textbf{Table} &\textbf{Text} & \textbf{Title}& \textbf{mAP$_{50}$} \\ 
        \hline
        \multirow{4}{*}{PLN$_4 \to $DLN$_4$}
        & \textcolor{lightgray}{Oracle} & \textcolor{lightgray}{80.77} & \textcolor{lightgray}{55.17} & \textcolor{lightgray}{74.02} & \textcolor{lightgray}{73.46} & \textcolor{lightgray}{70.85} \\
        & Source Only & 44.91 & 09.02 & 51.08 & 39.59 & 36.15 \\
        & IRG~\cite{10204606} & 46.51 & 09.02 & 54.71 & 42.15 & 38.10 \\
        \rowcolor{gray!20} & Ours & 46.62 & 07.49 & 62.25 & 45.08 & 40.36 \\
        \hline
        \end{tabular}
    }
\end{table*}

\vspace{-4ex}

Table~\ref{tab:dln10_m6doc10_style} illustrates adaptation from DocLayNet~\cite{10.1145/3534678.3539043} to M$^6$Doc~\cite{cheng2023M$^6$Doc}, a more challenging scenario due to high layout variability and multilingual content. Although IRG achieves a marginally higher mAP, our DLAdapter outperforms the model without adaptation, confirming that it captures some transferable features even under large domain shifts. The residual gap largely comes from the heterogeneous typography and fine-grained classes of M$^6$Doc~\cite{cheng2023M$^6$Doc}, which complicate pseudo-label generation in the absence of target supervision. Consequently, purely unsupervised alignment faces difficulty, especially for less frequent or domain-specific elements. Despite these obstacles, our DLAdapter dual-teacher design offers robust adaptability in the diverse dataset. 

\vspace{-4ex}

\begin{table*}[!ht]
    \centering
    \footnotesize
    \caption{Per-class AP and overall mAP$_{50}$ performance in domain adaptation scenario: \textbf{DLN$_{10}$} $\rightarrow$ \textbf{M$^6$Doc$_{10}$}.}
    \resizebox{\textwidth}{!}{
        \begin{tabular}{l|l|cccccccccc|c}
        \toprule
        \multirow{1}{*}{\textbf{Scenario}} &
        \multirow{1}{*}{\textbf{Method}} &
        \textbf{Caption} & 
        \textbf{Footnote} & 
        \textbf{Formula} & 
        \makecell{\textbf{Page-}\\\textbf{footer}} & 
        \makecell{\textbf{Page-}\\\textbf{header}} & 
        \textbf{Picture} & 
        \makecell{\textbf{Section-}\\\textbf{header}} & 
        \textbf{Table} & 
        \textbf{Text} & 
        \textbf{Title} & 
        \textbf{mAP$_{50}$} \\
        \midrule
        \multirow{4}{*}{DLN$_{10}$$\to$M$^6$Doc$_{10}$}
        & \textcolor{lightgray}{Oracle} & \textcolor{lightgray}{50.56} & \textcolor{lightgray}{66.86} & \textcolor{lightgray}{15.33} & \textcolor{lightgray}{89.12} & \textcolor{lightgray}{60.67} & \textcolor{lightgray}{78.39} & \textcolor{lightgray}{89.41} & \textcolor{lightgray}{64.09} & \textcolor{lightgray}{81.48} & \textcolor{lightgray}{79.94} & \textcolor{lightgray}{67.60} \\
        & Source Only & 24.82 & 04.71 & 02.53 & 13.56 & 00.24 & 40.19 & 01.85 & 23.71 & 36.24 & 9.05 & 15.69 \\
        & IRG~\cite{10204606} & 23.11 & 05.96 & 02.09 & 15.45 & 00.02 & 53.39 & 01.98 & 29.34 & 36.86 & 08.24 & 17.64 \\
        \rowcolor{gray!20} & Ours & 23.44 & 07.66 & 02.35 & 39.50 & 00.00 & 40.38 & 01.59 & 26.04 & 37.08 & 05.61 & 18.36 \\ 
        \hline
        \end{tabular}
    }
    \label{tab:dln10_m6doc10_style}
\end{table*}

\vspace{-4ex}

\subsubsection{Content-aware (Intra-dataset) Results.}
\label{subsubsec:intradataset}
We further evaluate source-free adaptation within DocLayNet~\cite{10.1145/3534678.3539043}, focusing on four different subcategories, \ie, \textit{financial reports}, \textit{scientific articles}, \textit{manuals}, and \textit{laws and regulations}. Table~\ref{tab:dln_intra_all} provides an overview of our DLAdapter performance compared to both a source-only baseline and the state-of-the-art SFOD approach, IRG~\cite{10204606}. As shown, our DLAdapter consistently outperforms IRG~\cite{10204606}, particularly on domain adaptation scenario \textit{Sci.$\rightarrow$Fin.}, where DLAdapter achieves $21.21\%$ at mAP$_{50}$, slightly above IRG~\cite{10204606} and significantly above the source-only performance. These improvements indicate that even when source and target domains belong to the same dataset family, distinct subcategory differences can pose considerable domain gaps, which our dual-teacher strategy effectively mitigates.

Table~\ref{tab:dln_intra_all} also provides a performance comparison for intra-dataset adaptation scenarios at the class level. In scenario \textit{Sci.$\rightarrow$Fin.}, IRG~\cite{10204606} exhibits stronger performance on \textit{List-item} and \textit{Picture}, whereas our method excels at \textit{Page-footer} and \textit{Table}. Similarly, on scenario \textit{Man.$\rightarrow$Fin.} and \textit{Laws.$\rightarrow$Man.}, our approach yields consistently high accuracy across classes, including \textit{Section-header} and \textit{Page-header}. Additionally, the Footnote class is scarce in the financial sub-domain, so even the Oracle reaches only 0.23 AP, demonstrating the intrinsic difficulty. Consequently, small numeric differences around this near-zero baseline are not statistically meaningful. Overall, these results reinforce that our DLAdapter framework can handle substantial sub-domain variations even within a single large dataset, reducing the reliance on labeled target data without sacrificing adaptation quality.

\begin{table*}[!ht]
    \centering
    \footnotesize
    \caption{Per-class AP and overall mAP$_{50}$ performance of intra-dataset domain adaptation scenarios on DocLayNet~\cite{10.1145/3534678.3539043}. Each block shows adaptation from one subcategory to another. There is no class \textit{Formula} in the subcategory \textit{Fin.} and no class \textit{Footnote} in the subcategory \textit{Man.} on DocLayNet~\cite{10.1145/3534678.3539043}.}
    \renewcommand{\arraystretch}{1.5}
    \resizebox{\textwidth}{!}{
        \begin{tabular}{l|l|ccccccccccc|c}
            \toprule
            \multirow{1}{*}{\textbf{Scenario}} & 
            \multirow{1}{*}{\textbf{Method}} & 
            \multicolumn{1}{c}{\textbf{Caption}} & 
            \multicolumn{1}{c}{\textbf{Footnote}} & 
            \multicolumn{1}{c}{\textbf{Formula}} & 
            \multicolumn{1}{c}{\textbf{List-item}} & 
            \makecell{\textbf{Page-}\\\textbf{footer}} & 
            \makecell{\textbf{Page-}\\\textbf{header}}& 
            \multicolumn{1}{c}{\textbf{Picture}} & 
            \makecell{\textbf{Section-}\\\textbf{header}}& 
            \multicolumn{1}{c}{\textbf{Table}} & 
            \multicolumn{1}{c}{\textbf{Text}} & 
            \multicolumn{1}{c}{\textbf{Title}} & 
            \multicolumn{1}{|c}{\textbf{mAP$_{50}$}} \\
            \midrule
            \multirow{4}{*}{\makecell{Sci. \\$\rightarrow$ Fin.}}
            & \textcolor{lightgray}{Oracle} & \textcolor{lightgray}{40.20} & \textcolor{lightgray}{00.23} & \textcolor{lightgray}{-} & \textcolor{lightgray}{58.23} & \textcolor{lightgray}{63.62} & \textcolor{lightgray}{51.86} & \textcolor{lightgray}{71.77} & \textcolor{lightgray}{46.00} & \textcolor{lightgray}{88.07} & \textcolor{lightgray}{84.71} & \textcolor{lightgray}{18.52} & \textcolor{lightgray}{52.32} \\
            & Source Only           & 00.28  & 00.23 & - & 39.63 & 06.20 & 11.64 & 12.86 & 18.33 & 47.38 & 50.57 & 00.04  & 18.72 \\
            & IRG~\cite{10204606}   & 00.46 & 00.43 & - & 41.69 & 06.43 & 10.36 & 16.92 & 21.69 & 55.72 & 54.40 & 00.11 & 20.82 \\
            \rowcolor{gray!20} & Ours & 00.29  & 00.50 & - & 41.47 & 06.68 & 10.60 & 15.93 & 21.30 & 58.10 & 57.19 & 0.03 & 21.21 \\
            \hline
            \multirow{4}{*}{\makecell{Man. \\$\rightarrow$ Fin.}}
            & \textcolor{lightgray}{Oracle} & \textcolor{lightgray}{40.20} & \textcolor{lightgray}{00.23} & \textcolor{lightgray}{-} & \textcolor{lightgray}{58.23} & \textcolor{lightgray}{63.62} & \textcolor{lightgray}{51.86} & \textcolor{lightgray}{71.77} & \textcolor{lightgray}{46.00} & \textcolor{lightgray}{88.07} & \textcolor{lightgray}{84.71} & \textcolor{lightgray}{18.52} & \textcolor{lightgray}{52.32} \\
            & Source Only           & 06.63  & 00.96 & - & 39.67 & 23.94 & 18.47 & 21.06 & 27.80 & 29.87 & 54.89 & 00.09 & 22.34 \\
            & IRG~\cite{10204606}   & 04.95  & 01.55 & - & 42.53 & 35.92 & 20.27 & 31.80 & 38.22 & 49.20 & 64.21 & 00.16 & 28.93 \\
            \rowcolor{gray!20} & Ours & 04.87  & 01.72 & - & 43.76 & 29.06 & 22.52 & 25.15 & 33.86 & 46.74 & 65.65 & 00.69 & 27.40 \\
            \hline
            \multirow{4}{*}{\makecell{Laws \\$\rightarrow$ Man.}}
            & \textcolor{lightgray}{Oracle} & \textcolor{lightgray}{79.51} & \textcolor{lightgray}{-} & \textcolor{lightgray}{00.00} & \textcolor{lightgray}{70.19} & \textcolor{lightgray}{72.08} & \textcolor{lightgray}{44.51} & \textcolor{lightgray}{96.70} & \textcolor{lightgray}{65.36} & \textcolor{lightgray}{49.07} & \textcolor{lightgray}{80.54} & \textcolor{lightgray}{57.10} & \textcolor{lightgray}{61.51} \\
            & Source Only           & 47.28 & -  & 00.00 & 61.19 & 57.44 & 94.25 & 87.16 & 46.54 & 26.21 & 72.15 & 01.12 & 49.33 \\
            & IRG~\cite{10204606}   & 41.86 & -  & 00.00 & 62.05 & 58.47 & 90.34 & 87.04 & 49.43 & 27.81 & 72.08 & 02.48 & 49.16 \\
            \rowcolor{gray!20} & Ours & 46.63 & -  & 00.00 & 57.16 & 59.69 & 96.24 & 87.23 & 50.40 & 29.64 & 69.09 & 02.46 & 49.85 \\
            \hline
        \end{tabular}
        \label{tab:dln_intra_all}
    }
\end{table*}

\subsection{Ablation Studies}
\label{subsec:ablation}
We conduct an ablation study on the \textbf{PLN$_4 \rightarrow$ DLN$_4$} scenario to assess the impact of each proposed component in our DLAdapter framework. Table~\ref{tab:ablation} lists the configurations in which certain modules are disabled (\ding{55}) or enabled (\ding{51}), showing their individual and combined contributions to the final performance (mAP$_{50}$). Specifically, we examine:
\begin{itemize}
  \setlength\itemsep{1em}
    \item \textbf{Hard Selection}: Using a one-size-fits-all threshold for pseudo labels selection;
    \item \textbf{Dynamic Selection}: Using the proposed consensus pseudo-labeling process;
    \item \textbf{Soft Label KL Distillation}: Computing the Kullback–Leibler divergence between the teacher model soft probability outputs and the student model predictions;
    \item \textbf{Auxiliary losses}: include distillation loss for knowledge transfer between teacher and student model, contrastive loss for enforcing consistency under different augmentations and entropy loss for confident predictions without source data.
\end{itemize}

As shown in Table~\ref{tab:ablation}, simply applying the \textbf{Source Only} model without any adaptation yields an initial mAP of $36.15\%$. Replacing it with \textbf{Hard Selection} alone improves performance to $37.90\%$. Combining Hard Selection with \textbf{Soft Label KL Distillation}  gives a similar but slightly lower score of $37.79\%$. By contrast, using \textbf{Dynamic Selection} alone achieves $38.54\%$, indicating that flexible pseudo-labeling generally outperforms fixed thresholds. Augmenting Dynamic Selection with Soft Label KL Distillation pushes the result to $39.36\%$. Finally, enabling all modules, including \textbf{Auxiliary Losses} for knowledge transfer, contrastive alignment, and entropy minimization, reaches the best mAP of \textbf{$40.36\%$}. These findings confirm that each component incrementally boosts performance and that our full configuration most effectively adapts the source model to the new domain.

\begin{table}[!htb]
    \centering
    \footnotesize
    \caption{Ablation study of DLAdapter on \textbf{PLN$_4 \rightarrow$ DLN$_4$}. 
    Each row corresponds to activating or deactivating specific components/settings of our DLAdapter framework. The final configuration achieves the best mAP$_{50}$.}
    \label{tab:ablation}
    \renewcommand{\arraystretch}{1.2}
    \resizebox{\textwidth}{!}{
        \begin{tabular}{c|c|c|c|c|c}
        \toprule
        \textbf{Source Only} & 
        \textbf{\makecell{Hard Selection}} & 
        \textbf{\makecell{Dynamic Selection}} & 
        \textbf{\makecell{Soft Label\\KL Distillation}} & 
        \textbf{\makecell{Auxiliary Losses}}& 
        \textbf{mAP$_{50}$} \\
        \midrule
        \ding{51} & \ding{55} & \ding{55} & \ding{55} & \ding{55} & 36.15 \\
        \ding{55} & \ding{51} & \ding{55} & \ding{55} & \ding{55} & 37.90 \\
        \ding{55} & \ding{51} & \ding{55} & \ding{51} & \ding{55} & 37.79 \\
        \ding{55} & \ding{55} & \ding{51} & \ding{55} & \ding{55} & 38.54 \\
        \ding{55} & \ding{55} & \ding{51} & \ding{51} & \ding{55} & 39.36 \\
        \rowcolor{gray!20} \ding{55} & \ding{55} & \ding{51} & \ding{51} & \ding{51} & 40.36 \\
        \hline
        \end{tabular}
    }
\end{table}

\subsection{Qualitative Analysis}
\label{subsec:qualitative}

\begin{figure*}[htbp]
    \centering
    \includegraphics[width=0.98\textwidth]{Images/visualization_14757.png}
    \caption{A qualitative visualization of our adaptation process with DocLayNet \cite{10.1145/3534678.3539043}. The left panel shows ground-truth annotations, while the middle and right panels display the pseudo-labels produced by DLAdapter after the 1st and 4th adaptation epoch.}
    \label{fig:qualitative}
\end{figure*}

In addition to quantitative metrics, we present a qualitative assessment of our DLAdapter framework on the DocLayNet dataset~\cite{10.1145/3534678.3539043}. Figure~\ref{fig:qualitative} shows an example page with its ground-truth annotations (right) and the pseudo labels predicted by our method after 1 and 4 training epochs. Visually, the detection quality improves substantially over time, with more accurate bounding boxes and fewer misclassifications by the fourth epoch. Nevertheless, these results also clearly show the limitations of this method. If the pseudo labels were perfect and thus the source model, we would not need domain adaptation. This also means that the existence of the target labels still provides great added value and that without them, there is a significant loss in performance.

%% file: chapters/conclusion_and_future_work.tex
\section{Conclusion}

In this paper, we introduced \textbf{DLAdapter}, the first method that integrates Document Layout Analysis (DLA) with source-free domain adaptation under the constraint that both target labels and source data are inaccessible. Through extensive experiments across multiple datasets, DLAdapter consistently outperforms source-only baselines and achieves state-of-the-art performance among existing SFDA methods. Nonetheless, there remains a gap compared to fully supervised target-domain models, largely due to significant domain discrepancies. We hope that this work will stimulate further research in both the SFDA and DLA communities, fostering more robust and privacy-preserving document analysis.

\subsubsection{Limitations} Despite its promising results, DLAdapter is adapted without labeled target data, and errors in pseudo-labels can accumulate under large domain gaps. Our current approach focuses mainly on visual input, while textual input with OCR could improve adaptation for semantically driven layout elements. The dual-teacher design, though effective, adds extra training and inference overhead. Addressing these issues could lead to more robust and efficient source-free domain adaptation in document layout analysis, enabling broader real-world applicability.

\subsubsection{Acknowledgment} This work was supported in part by Helmholtz Association of German Research Centers, in part by the Ministry of Science, Research and the Arts of Baden-Württemberg (MWK) through the Cooperative Graduate School Accessibility through AI-based Assistive Technology (KATE) under Grant BW6-03, and in part by Karlsruhe House of Young Scientists (KHYS). This work was partially performed on the HoreKa supercomputer funded by the MWK and by the Federal Ministry of Education and Research, partially on the HAICORE@KIT partition supported by the Helmholtz Association Initiative and Networking Fund, and partially on bwForCluster Helix supported by the state of Baden-Württemberg through bwHPC and the German Research Foundation (DFG) through grant INST 35/1597-1 FUGG.

%% file: main.bbl
\begin{thebibliography}{10}
\providecommand{\url}[1]{\texttt{#1}}
\providecommand{\urlprefix}{URL }
\providecommand{\doi}[1]{https://doi.org/#1}

\bibitem{bhatt2021survey}
Bhatt, J., Hashmi, K.A., Afzal, M.Z., Stricker, D.: A survey of graphical page object detection with deep neural networks. Applied Sciences  \textbf{11}(12), ~5344 (2021)

\bibitem{bhowmik2023document}
Bhowmik, S.: Document layout analysis. Springer (2023)

\bibitem{chen2022learningdomainadaptiveobject}
Chen, M., Chen, W., Yang, S., Song, J., Wang, X., Zhang, L., Yan, Y., Qi, D., Zhuang, Y., Xie, D., Pu, S.: Learning domain adaptive object detection with probabilistic teacher (2022)

\bibitem{chen2024rodla}
Chen, Y., Zhang, J., Peng, K., Zheng, J., Liu, R., Torr, P., Stiefelhagen, R.: Rodla: Benchmarking the robustness of document layout analysis models (2024)

\bibitem{10.1145/3581783.3612273}
Chen, Z., Wang, Z., Zhang, Y.: Exploiting low-confidence pseudo-labels for source-free object detection. In: Proceedings of the 31st ACM International Conference on Multimedia. p. 5370–5379. MM '23, Association for Computing Machinery, New York, NY, USA (2023)

\bibitem{cheng2024towards}
Cheng, C., Yue, Z., Zhang, H.: Towards debiased source-free domain adaptation (2024)

\bibitem{cheng2023M$^6$Doc}
Cheng, H., Zhang, P., Wu, S., Zhang, J., Zhu, Q., Xie, Z., Li, J., Ding, K., Jin, L.: M$^{6}$doc: A large-scale multi-format, multi-type, multi-layout, multi-language, multi-annotation category dataset for modern document layout analysis (2023)

\bibitem{10.1609/aaai.v37i1.25119}
Chu, Q., Li, S., Chen, G., Li, K., Li, X.: Adversarial alignment for source free object detection. In: Proceedings of the Thirty-Seventh AAAI Conference on Artificial Intelligence and Thirty-Fifth Conference on Innovative Applications of Artificial Intelligence and Thirteenth Symposium on Educational Advances in Artificial Intelligence. AAAI'23/IAAI'23/EAAI'23, AAAI Press (2023)

\bibitem{da2023vision}
Da, C., Luo, C., Zheng, Q., Yao, C.: Vision grid transformer for document layout analysis. In: Proceedings of the IEEE/CVF international conference on computer vision. pp. 19462--19472 (2023)

\bibitem{5206848}
Deng, J., Dong, W., Socher, R., Li, L.J., Li, K., Fei-Fei, L.: Imagenet: A large-scale hierarchical image database. In: 2009 IEEE Conference on Computer Vision and Pattern Recognition. pp. 248--255 (2009)

\bibitem{deng2024yolo}
Deng, Q., Ibrayim, M., Hamdulla, A., Zhang, C.: The yolo model that still excels in document layout analysis. Signal, Image and Video Processing  \textbf{18}(2),  1539--1548 (2024)

\bibitem{7780459}
He, K., Zhang, X., Ren, S., Sun, J.: Deep residual learning for image recognition. In: 2016 IEEE Conference on Computer Vision and Pattern Recognition (CVPR). pp. 770--778 (2016)

\bibitem{huang2022layoutlmv3}
Huang, Y., Lv, T., Cui, L., Lu, Y., Wei, F.: Layoutlmv3: Pre-training for document ai with unified text and image masking (2022)

\bibitem{khanh2024dynamicretrainingupdatingmeanteacher}
Khanh, T.L.B., Nguyen, H.H., Pham, L.H., Tran, D.N.N., Jeon, J.W.: Dynamic retraining-updating mean teacher for source-free object detection (2024)

\bibitem{9528982}
Kim, Y., Cho, D., Han, K., Panda, P., Hong, S.: Domain adaptation without source data. IEEE Transactions on Artificial Intelligence  \textbf{2}(6),  508--518 (2021)

\bibitem{kothandaraman2023salad}
Kothandaraman, D., Shekhar, S., Sancheti, A., Ghuhan, M., Shukla, T., Manocha, D.: Salad: Source-free active label-agnostic domain adaptation for classification, segmentation and detection. In: Proceedings of the IEEE/CVF Winter Conference on Applications of Computer Vision. pp. 382--391 (2023)

\bibitem{kundu2020universal}
Kundu, J.N., Venkat, N., Babu, R.V., et~al.: Universal source-free domain adaptation. In: Proceedings of the IEEE/CVF conference on computer vision and pattern recognition. pp. 4544--4553 (2020)

\bibitem{li2024daadalearningdomainawareadapter}
Li, H., Zhang, R., Yao, H., Zhang, X., Hao, Y., Song, X., Li, X., Zhao, Y., Li, L., Chen, Y.: Da-ada: Learning domain-aware adapter for domain adaptive object detection (2024)

\bibitem{10452835}
Li, J., Yu, Z., Du, Z., Zhu, L., Shen, H.T.: A comprehensive survey on source-free domain adaptation. IEEE Transactions on Pattern Analysis and Machine Intelligence pp. 1--22 (2024)

\bibitem{li2024comprehensive}
Li, J., Yu, Z., Du, Z., Zhu, L., Shen, H.T.: A comprehensive survey on source-free domain adaptation. IEEE Transactions on Pattern Analysis and Machine Intelligence  (2024)

\bibitem{10.1145/3503161.3547911}
Li, J., Xu, Y., Lv, T., Cui, L., Zhang, C., Wei, F.: Dit: Self-supervised pre-training for document image transformer. In: Proceedings of the 30th ACM International Conference on Multimedia. p. 3530–3539. MM '22, Association for Computing Machinery, New York, NY, USA (2022)

\bibitem{li2020cross}
Li, K., Wigington, C., Tensmeyer, C., Zhao, H., Barmpalios, N., Morariu, V.I., Manjunatha, V., Sun, T., Fu, Y.: Cross-domain document object detection: Benchmark suite and method. In: Proceedings of the IEEE/CVF Conference on Computer Vision and Pattern Recognition. pp. 12915--12924 (2020)

\bibitem{9878540}
Li, S., Ye, M., Zhu, X., Zhou, L., Xiong, L.: Source-free object detection by learning to overlook domain style. In: 2022 IEEE/CVF Conference on Computer Vision and Pattern Recognition (CVPR). pp. 8004--8013 (2022)

\bibitem{Li2020AFL}
Li, X., Chen, W., Xie, D., Yang, S., Yuan, P., Pu, S., Zhuang, Y.: A free lunch for unsupervised domain adaptive object detection without source data. Proceedings of the AAAI Conference on Artificial Intelligence, 35  \textbf{AAAI Technical Track on Machine Learning III} (2020)

\bibitem{10377810}
Liu, Q., Lin, L., Shen, Z., Yang, Z.: Periodically exchange teacher-student for source-free object detection. In: 2023 IEEE/CVF International Conference on Computer Vision (ICCV). pp. 6391--6401. IEEE Computer Society, Los Alamitos, CA, USA (Oct 2023)

\bibitem{liu2023PETS}
Liu, Q., Lin, L., Shen, Z., Yang, Z.: Periodically exchange teacher-student for source-free object detection. In: Proceedings of the IEEE/CVF International Conference on Computer Vision. pp. 6414--6424 (2023)

\bibitem{liu2021source}
Liu, Y., Zhang, W., Wang, J.: Source-free domain adaptation for semantic segmentation. In: Proceedings of the IEEE/CVF conference on computer vision and pattern recognition. pp. 1215--1224 (2021)

\bibitem{mohamadi2024featurebasedmethodsdomain}
Mohamadi, H., Keyvanrad, M.A., Mohammadi, M.R.: Feature based methods domain adaptation for object detection: A review paper (2024)

\bibitem{9157725}
Nath~Kundu, J., Venkat, N., Rahul, M.V., Venkatesh~Babu, R.: Universal source-free domain adaptation. In: 2020 IEEE/CVF Conference on Computer Vision and Pattern Recognition (CVPR). pp. 4543--4552 (2020)

\bibitem{10.1145/3534678.3539043}
Pfitzmann, B., Auer, C., Dolfi, M., Nassar, A.S., Staar, P.: Doclaynet: A large human-annotated dataset for document-layout segmentation. In: Proceedings of the 28th ACM SIGKDD Conference on Knowledge Discovery and Data Mining. p. 3743–3751. KDD '22, Association for Computing Machinery, New York, NY, USA (2022)

\bibitem{NIPS2015_14bfa6bb}
Ren, S., He, K., Girshick, R., Sun, J.: Faster r-cnn: Towards real-time object detection with region proposal networks. In: Cortes, C., Lawrence, N., Lee, D., Sugiyama, M., Garnett, R. (eds.) Advances in Neural Information Processing Systems. vol.~28. Curran Associates, Inc. (2015)

\bibitem{10.5555/3294771.3294885}
Tarvainen, A., Valpola, H.: Mean teachers are better role models: Weight-averaged consistency targets improve semi-supervised deep learning results. In: Proceedings of the 31st International Conference on Neural Information Processing Systems. p. 1195–1204. NIPS'17, Curran Associates Inc., Red Hook, NY, USA (2017)

\bibitem{10204606}
VS, V., Oza, P., Patel, V.M.: Instance relation graph guided source-free domain adaptive object detection. In: 2023 IEEE/CVF Conference on Computer Vision and Pattern Recognition (CVPR). pp. 3520--3530. IEEE Computer Society, Los Alamitos, CA, USA (Jun 2023)

\bibitem{vs2023instance}
VS, V., Oza, P., Patel, V.M.: Instance relation graph guided source-free domain adaptive object detection. In: Proceedings of the IEEE/CVF conference on computer vision and pattern recognition. pp. 3520--3530 (2023)

\bibitem{wu2019detectron2}
Wu, Y., Kirillov, A., Massa, F., Lo, W.Y., Girshick, R.: Detectron2. \url{https://github.com/facebookresearch/detectron2} (2019)

\bibitem{xia2021adaptive}
Xia, H., Zhao, H., Ding, Z.: Adaptive adversarial network for source-free domain adaptation. In: Proceedings of the IEEE/CVF international conference on computer vision. pp. 9010--9019 (2021)

\bibitem{XIAO2024110533}
Xiao, Y., Xiao, G., Chen, H.: Unified multi-level neighbor clustering for source-free unsupervised domain adaptation. Pattern Recognition  \textbf{153},  110533 (2024)

\bibitem{xiong2022source}
Xiong, L., Ye, M., Zhang, D., Gan, Y., Liu, Y.: Source data-free domain adaptation for a faster r-cnn. Pattern Recognition  \textbf{124},  108436 (2022)

\bibitem{yoon2024enhancingsourcefreedomainadaptive}
Yoon, I., Kwon, H., Kim, J., Park, J., Jang, H., Sohn, K.: Enhancing source-free domain adaptive object detection with low-confidence pseudo label distillation (2024)

\bibitem{10096635}
Zhang, S., Zhang, L., Liu, Z.: Refined pseudo labeling for source-free domain adaptive object detection. In: ICASSP 2023 - 2023 IEEE International Conference on Acoustics, Speech and Signal Processing (ICASSP). pp.~1--5 (2023)

\bibitem{zhong2019publaynet}
Zhong, X., Tang, J., Yepes, A.J.: Publaynet: largest dataset ever for document layout analysis (2019)

\end{thebibliography}
